\documentclass{article}
\usepackage{spconf,amsmath,graphicx}
\usepackage{caption}
\usepackage{subcaption}
\usepackage{mathrsfs}
\usepackage{braket}
\usepackage{multirow}
\usepackage{url}
\usepackage[numbers,sort&compress]{natbib}

\title{End-to-End Joint Learning of Natural Language Understanding and Dialogue Manager}

\name{Xuesong Yang$^\star$, Yun-Nung Chen$^{\S}$, Dilek Hakkani-T\"{u}r$^\dag$, Paul Crook$^\ast$, Xiujun Li$^\ddag$, Jianfeng Gao$^\ddag$, Li Deng$^\ddag$}

\address{$^\star$University of Illinois at Urbana-Champaign, Urbana, IL, USA \\$^\S$National Taiwan University, Taipei, Taiwan\\$\dag$Google Research, Mountain View, CA, USA\\ $^\ddag$Microsoft Research, Redmond, WA, USA,\quad $^\ast$Microsoft Corporation, Redmond, WA, USA}
\begin{document}
%
\maketitle
\begin{abstract}
Natural language understanding and dialogue policy learning are both essential in conversational systems that predict the next system actions in response to a current user utterance. Conventional approaches aggregate separate models of natural language understanding (NLU) and system action prediction (SAP) as a pipeline that is sensitive to noisy outputs of error-prone NLU\@. To address the issues, we propose an end-to-end deep recurrent neural network with limited contextual dialogue memory by jointly training NLU and SAP on DSTC4 multi-domain human-human dialogues. Experiments show that our proposed model significantly outperforms the state-of-the-art pipeline models for both NLU and SAP, which indicates that our joint model is capable of mitigating the affects of noisy NLU outputs, and NLU model can be refined by error flows backpropagating from the extra supervised signals of system actions.
\end{abstract}
%



\section{Introduction}\label{sec:intro}
%
%

Recent progress of designing conversational agents for commercial purposes, such as Microsoft's Cortana, Apple's Siri, and Amazon's Echo, has attracted more attention from both academia and industry. Two essential components of these conversational agents are natural language understanding~(NLU) and dialog manager~(DM). NLU typically detects dialog domains by parsing user utterances followed by user intent classification and filling associated slots according to a domain-specific semantic template~\cite{tur2011spoken}; DM keeps monitoring the belief distribution over all possible user states underlying current user behaviors, and predicts responsive system actions~\cite{young2013pomdp,dhingra2016end}. For example, given a user utterance ``\emph{any action movies recommended this weekend?}'', NLU predicts intent $\mathtt{request\_movie}$ and slots $\mathtt{genre}$ and $\mathtt{date}$; thereafter, DM predicts system action $\mathtt{request\_location}$. 

Traditional approaches for NLU usually model tasks of domain/intent classification and slot filling separately. Sequential labeling methods, such as hidden Markov models (HMMs) and conditional random field (CRF) are widely used in slot tagging tasks~\cite{wang2005spoken,raymond2007generative,lafferty2001conditional}; maximum entropy and support vector machines with linear kernel (LinearSVM) are applied to user intent prediction~\cite{haffner2003optimizing,yang2014machine,chelba2003speech}. These models highly rely on careful feature engineering that is laborious and time-consuming. Deep learning techniques making incredible progress on learning expressive feature representations have achieved better solutions to NLU modeling in ATIS domain~\cite{sarikaya2011deep,sarikaya2014application,wang2016learning,chen2016knowledge,chen2016syntax}. The performance was improved significantly by incorporating recurrent neural networks (RNN) and CRF model~\cite{yao2014spoken,mesnil2015using,yao2014recurrent}. Convolutional neural networks are also used for domain/intent classification~\cite{xu2014contextual,tur2012towards}. 

Slot tags and intents, as semantics representations of user behaviors, may share knowledge with each other such that separate modeling of these two tasks is constrained to take full advantage of all supervised signals. Flexible architectures of neural networks provide a way of jointly training with intent classification and slot filling~\cite{bhargava2013easy,xu2013convolutional}. Contextual information of previous queries and domain/intent prediction was also incorporated into RNN structures~\cite{shi2015contextual,chen2016end2end}.

Information flows from NLU to DM, and noisy outputs of NLU are apt to transfer errors to the following DM, so that it brings in challenges for monitoring the belief distribution and predicting system actions. Most successful approaches cast the dialog manager as a partially observable Markov decision process~\cite{young2013pomdp}, which uses hand-crafted features to represent the state and action space, and requires a large number of annotated conversations~\cite{henderson2014word} or human interactions~\cite{gavsic2013line,wen2016network}. Converting these methods into practice is far from trivial, and exact policy learning is computational intractable. Therefore, they are constrained to narrow domains. 

In order to address the above problems, we propose an end-to-end deep RNN with limited contextual dialog memory that can be jointly trained by three supervised signals---user slot tagging, intent prediction and system action prediction (SAP). Our model expresses superb advantages in natural language understanding and dialog manager. Highly expressive feature representations beyond conventional aggregation of slot tags and intents are expected to be captured in our joint model, so that the affects of noisy output from NLU can be mitigated. Extra supervised signal from system actions is capable of refining NLU model by backpropagating the associated error gradients.

\section{End-to-End Joint Model}\label{sec:proposed_models}

The joint model can be considered as a SAP model stacked on top of a history of NLU models (see Fig.~\ref{fig:joint_blstm}).
NLU model is designed as a multi-tasking framework by sharing bi-directional long short-term memory (biLSTM) layers with slot tagging and intent prediction.

\subsection{Sequence to Sequence Model with biLSTM Cells}\label{sec:bilstm}
Given a sequence of input vectors $\mathbf{x}$=$\set{x_t}_{1}^{T}$, a recurrent unit $\mathcal{H}$ computes a sequence of hidden vectors $\mathbf{h}$=$\set{h_t}_{1}^{T}$ and a sequence of output symbols $\mathbf{\hat{y}}$=$\set{\hat{y}_t}_{1}^{T}$ by iterating the following equations,
\small
\begin{align}
	h_t &= \mathcal{H}\left(x_t, h_{t-1}\right) = \sigma\left(W_{xh} x_t +  U_{hh} h_{t-1}\right) \nonumber\\
	\hat{y_t} &= \arg\max \left(\mathtt{softmax}\left(W_{hy} h_t\right) \right) \nonumber
\end{align}
\normalsize
where $\mathtt{softmax}\left(z_m\right)=e^{z_m} / \sum_{i} e^{z_i}$, $\sigma$ is an activation function, and $W_{xh}$, $U_{hh}$ and $W_{hy}$ are weight matrices. The goal of sequence to sequence model (Seq2Seq) is to estimate a conditional probability $p\left(\mathbf{\hat{y}} | \mathbf{x}\right) = \prod_{t=1}^{T} p\left(\hat{y}_t | \mathbf{x}\right)$
such that the distance (loss) between predicted distribution $p\left(\hat{y_t}|\mathbf{x}\right)$ and target distribution $q\left(y_t|\mathbf{x}\right)$ is minimized, namely,
\small
\begin{equation*}
	\mathtt{loss} = - \sum_{t=1}^{T} \sum_{z=1}^{M} q\left(y_t=z|\mathbf{x}\right) \log p\left(\hat{y_t}=z|\mathbf{x}\right)
\end{equation*}
\normalsize
where $M$ is the number of unique output labels. The loss of this Seq2Seq model can be optimized using backpropagation. LSTM cells are chosen as recurrent units since LSTM can mitigate problems of vanishing or exploding gradients in long-term dependencies via self-regularization~\cite{hochreiter1997long}. The LSTM recurrent unit $\mathcal{H}$ can be further expanded as,
\small
\begin{align}
	h_t &= \mathcal{H}\left(x_t, h_{t-1}\right) = o_t \odot \mathtt{tanh}\left(c_t\right) \nonumber\\
	c_t &= f_t \odot c_{t-1} + i_t \odot g_t \nonumber\\
	o_t &= \mathtt{sigm}\left(W_{xo} x_t + U_{ho} h_{t-1}\right) \nonumber,\; i_t = \mathtt{sigm}\left(W_{xi} x_t + U_{hi} h_{t-1}\right) \nonumber\\
	f_t &= \mathtt{sigm}\left(W_{xf} x_t + U_{hf} h_{t-1}\right) \nonumber,\; g_t = \mathtt{tanh}\left(W_{xg} x_t + U_{hg} h_{t-1}\right) \nonumber
\end{align}
\normalsize
where the sigmoid functions $\mathtt{sigm}$ and $\mathtt{tanh}$ are applied element-wise, and $\odot$ denotes element-wise product. Since preceding and following lexical contexts are important in analysis of user utterances, bi-directional LSTM cells~\cite{schuster1997bidirectional} are used. Therefore sequence $\mathbf{x}$ and its reverse go through LSTM layers separately, followed by the concatination of the corresponding forward output $\overrightarrow{\mathbf{h}}$ and backward output $\overleftarrow{\mathbf{h}}$, 
\small
\begin{align}
	\overrightarrow{h}_t &= \mathcal{H}\left(x_t, \overrightarrow{h}_{t-1}\right), \quad \overleftarrow{h}_t = \mathcal{H}\left(x_t, \overleftarrow{h}_{t+1}\right) \nonumber\\
	\hat{y_t} &= \arg\max \left(\mathtt{softmax}\left(\overrightarrow{W}_{hy}\overrightarrow{h}_t + \overleftarrow{W}_{hy}\overleftarrow{h}_t\right)\right) \nonumber
\end{align}
\normalsize
where $\overrightarrow{W}_{hy}$ and $\overleftarrow{W}_{hy}$ are bi-directional weight matrices.

\subsection{Joint Modeling}
\begin{figure}[htbp]
\vspace{-2mm}
\includegraphics[width=.95\linewidth]{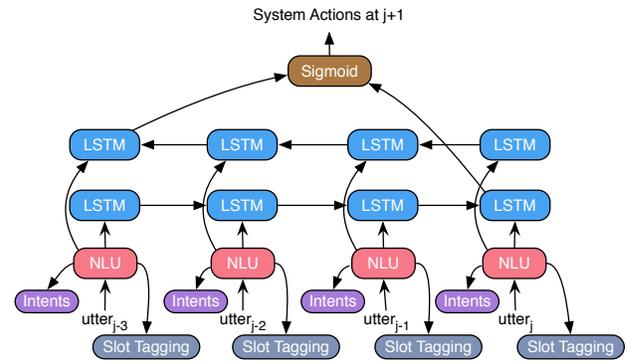}
\caption{Proposed end-to-end joint model}
\label{fig:joint_blstm}
\end{figure}

\begin{figure}[htbp]
\includegraphics[width=0.95\linewidth]{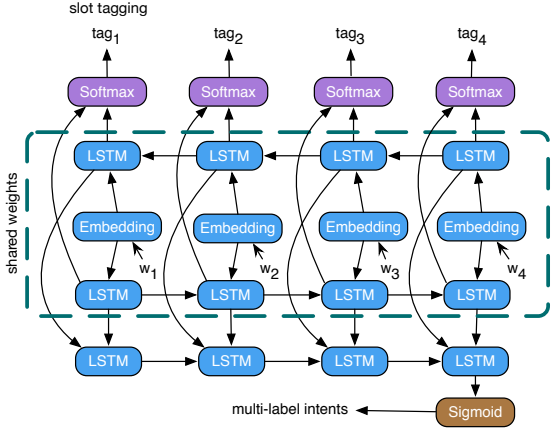}
\caption{biLSTMs-based NLU model}
\label{fig:nlu_blstm}
\vspace{-3mm}
\end{figure}

The proposed joint model is a RNN classifier that utilizes bi-directional LSTM cells $\mathcal{H}$, which takes as inputs $I$-$1$ history of current hidden outputs $\mathbf{h}_j^{(nlu)}=\set{h_i^{(nlu)}}_{j-I+1}^{j}$ from NLU units and performs one-vs-all binary classifications for SAP at the output layer (see Fig.~\ref{fig:joint_blstm}), in other word,
\small
\begin{align}
	\overrightarrow{h}_{i}^{(act)} &= \mathcal{H}\left(h_{i}^{(nlu)}, \overrightarrow{h}_{i-1}^{(act)}\right), \quad \overleftarrow{h}_{i}^{(act)} = \mathcal{H}\left(h_{i}^{(nlu)}, \overleftarrow{h}_{i+1}^{(act)}\right) \nonumber\\
	p^{(act)} &= \mathtt{sigm}\left(\overrightarrow{W}_{hy}^{(act)} \overrightarrow{h}_{j}^{(act)} + \overleftarrow{W}_{hy}^{(act)} \overleftarrow{h}_{j}^{(act)}\right) \nonumber \\
	\hat{y}_{k}^{(act)} &= \left\{\begin{matrix}
					     1, & p_{k}^{(act)} \ge \mathtt{threshold}\\ 
						 0, & otherwise 
	\end{matrix}\right. \nonumber
\end{align}
\normalsize
where $k \in \left[1, K\right]$ denotes the index of system action labels. 
NLU model at the $i$-th history is considered as a multi-task joint model with shared biLSTM layers for two tasks, where it takes as inputs a sequence of word vectors $\mathbf{w}=\set{w_t}_{1}^{T}$, and performs Seq2Seq for slot tagging and one-vs-all binary classifications for intent prediction (see Fig.~\ref{fig:nlu_blstm}). The biLSTM architecture mentioned in Section~\ref{sec:bilstm} can be directly applied to slot tagging task with $M$ unique user slot tags,
\small
\begin{align}
	\overrightarrow{h}_{t}^{1(i)} &= \mathcal{H}\left(w_t, \overrightarrow{h}_{t-1}^{1(i)}\right), \quad \overleftarrow{h}_{t}^{1(i)} = \mathcal{H}\left(w_t, \overleftarrow{h}_{t+1}^{1(i)}\right) \nonumber\\
	\hat{y}_{t}^{(tag_i)} &= \arg \max \left( \mathtt{softmax}\left(\overrightarrow{W}_{hy}^{(tag)}\overrightarrow{h}_{t}^{1(i)} + \overleftarrow{W}_{hy}^{(tag)}\overleftarrow{h}_{t}^{1(i)}\right) \right) \nonumber
\end{align}
\normalsize
where $\overrightarrow{h}_{t}^{1(i)}$ and $\overleftarrow{h}_{t}^{1(i)}$ denotes hidden outputs of the shared forward and backward layers, respectively. 
As for intent prediction task in NLU, we add one more recurrent LSTM layer on top of biLSTM layers, and only consider the last hidden vector $h_{T}^{2(int_i)}$ as the output of this second recurrent layer. Real human-human dialogs encode various number of intents in a single user utterance, and therefore, we design a set of one-vs-all binary classifiers at the output layer where each neuron is activated using a sigmoid function. The positive label of each classifier is predicted if its probability is no less than the threshold,
\small
\begin{align}
	h_{t}^{2(int_i)} &= \mathcal{H}\left(h_{t-1}^{2(int_i)}, \overrightarrow{h}_{t}^{1(i)}, \overleftarrow{h}_{t}^{1(i)}\right) \nonumber \\
	p^{(int_i)} &= \mathtt{sigm}\left(W_{hy}^{2(int)} h_{T}^{2(int_i)}\right) \nonumber\\
	\hat{y}_{n}^{(int_i)} &= \left\{\begin{matrix}
					     1, & p_{n}^{(int_i)} \ge \mathtt{threshold} \nonumber\\ 
						 0, & otherwise 
\end{matrix}\right.
\end{align}
\normalsize
where $n \in \left[1, N\right]$ is the index of intent labels. We choose the same two-layer recurrent architecture as the intent model to calculate the hidden vector $h_{i}^{(nlu)}$ out from the $i$-th NLU component with the size of $M$+$N$, where $M$ and $N$ are the number of unique slot tags and unique intents, respectively. 
\small
\begin{align}
	h_{i}^{(nlu)} = h_{T}^{2(nlu_i)}, \quad h_{t}^{2(nlu_i)} = \mathcal{H}\left(h_{t-1}^{2(nlu_i)}, \overrightarrow{h}_{t}^{1(i)}, \overleftarrow{h}_{t}^{1(i)}\right) \nonumber
\end{align}
\normalsize

End-to-end joint training estimates the conditional probability given a history of word vectors $\mathbf{w}_h = \set{\mathbf{w}^{(i)}}_{1}^{I}$ such that $\mathtt{loss} = l^{(act)} + l^{(tag)} + l^{(int)}$ is minimized, where
\footnotesize
\begin{align}
l^{(act)} &= -\sum_{k=1}^{K} \sum_{z=0}^{1}q\left(y_{k}^{(act)} = z \mid \mathbf{w}_h\right) \log p\left(\hat{y}_{k}^{(act)} = z \mid \mathbf{w}_h\right)\nonumber\\
l^{(tag)} &= - \sum_{i=1}^{I}\sum_{t=1}^{T} \sum_{z=1}^{M} q\left(y_{t}^{(tag_i)}=z|\mathbf{w}^{(i)}\right) \log p\left(\hat{y}_{t}^{(tag_i)}=z|\mathbf{w}^{(i)}\right)\nonumber \\
l^{(int)} &= -\sum_{i=1}^{I}\sum_{n=1}^{N}\sum_{z=0}^{1}q\left(y_{n}^{(int_i)} = z \mid \mathbf{w}^{(i)}\right) \log p\left(\hat{y}_{n}^{(int_i)} = z \mid \mathbf{w}^{(i)}\right) \nonumber 
\end{align}
\normalsize

\section{Experiments}
\label{sec:experiments}

\subsection{Training Configurations}
We choose a mini-batch stochastic gradient descent method Adam~\cite{kingma2014adam} with the batch size of 32 examples. The size of each hidden recurrent layer is 256, and the size of hidden output vector of NLU units is $M$+$N$, where $M$ and $N$ are the size of unique slot tags and intents, respectively. We assume the joint model can only get access to previous history with $I$=$5$. The dimension of word embeddings is 512. Dropout rate is 0.5. We apply 300 training epochs without using any early stop strategy. 
Best models for three tasks are selected separately upon decision thresholds well tuned on dev set under different metrics. Token-level micro-average F1 score is used for slot filling; frame-level accuracy (it counts only when the whole frame parse is correct) is used for user intent prediction and system action prediction. 
The code is released\footnote{\scriptsize\url{https://github.com/XuesongYang/end2end_dialog.git}}.
\begin{table}[htbp]
	\centering
	\caption{\emph{Statistics of data used in experiments. '\#' represents the number of unique items.}}\label{tab:dstc}
	\vspace{-2mm}
	\begin{tabular}{|c|ccccc|}
    \hline
              & \#utters & \#words & \#tags & \#intents & \#actions \\
    \hline\hline
		train &    5,648 &  2,252  &  87 &  68 & 66 \\
		dev   &    1,939 &  1,367  &  79 &  54 & 53 \\
		test  &    3,178 &  1,752  &  75 &  58 & 58 \\
    \hline
	\end{tabular}
\end{table}
\vspace{-3mm}
\begin{table}[htbp]
	\centering
	\caption{\emph{Performance (\%) of end-to-end models for SAP\@. F1, P, R are micro-averaged token-level scores; FrmAcc is frame-level accuracy. Oracle models are provided as references.}}\label{tab:joint}
	\vspace{-2mm}
	\begin{tabular}{|l|rrr|r|}
\hline
		Models & \multicolumn{1}{c}{F1} & \multicolumn{1}{c}{P} & \multicolumn{1}{c}{R} & \multicolumn{1}{|c|}{FrmAcc} \\
\hline\hline
		Baseline (CRF+SVMs) & 31.15 & 29.92 & 32.48	& 7.71 \\
		Pipeline (biLSTMs) & 19.89	& 14.87 & 30.01 & 11.96\\
		JointModel & 19.04 & 18.53 & 19.57 & \textbf{22.84}\\
\hline
		Oracle-SAP (SVMs) & 30.61 & 30.20 & 31.04 & 7.65 \\
		Oracle-SAP (biLSTM) & 23.09 & 22.24 & 24.01 & 19.67 \\
\hline
	\end{tabular}
\end{table}
\begin{table*}[htbp]
\centering
\caption{\emph{Performance (\%) of NLU models, where F1, Precision and Recall are at token-level and FrmAcc is at frame-level.}}
\vspace{-2mm}
\label{tab:nlu}
\begin{tabular}{|l|cccc|cccc|c|}
\hline
\multicolumn{1}{|c|}{\multirow{2}{*}{Models}} & \multicolumn{4}{c|}{User Slot Tagging (UST)} & \multicolumn{4}{c|}{User Intent Prediction (UIP)} & NLU (UST+UIP)  \\
\cline{2-10}
& F1	& Precision & Recall & FrmAcc & F1 & Precision & Recall & FrmAcc & FrmAcc\\
\hline\hline
NLU-Baseline & 40.50 & 61.41 & 30.21 & \textbf{77.31} & 49.75 & 52.56 & 47.24 & 37.19 & 33.13\\
NLU-Pipeline & 46.15 & 54.63 & 39.96 & 76.84 & 47.48 & 52.19 & 43.55 & 39.96 & 36.38\\
NLU-JointModel & 45.04 & 53.35	& 38.97 & 76.49 & 49.67 & 52.22 & 47.35 & \textbf{42.20} & \textbf{37.38}\\
\hline
\end{tabular}
\end{table*}

\vspace{-4mm}
\subsection{Corpus}
DSTC4 corpus\footnote{\scriptsize\url{http://www.colips.org/workshop/dstc4/data.html}} is selected, which collected human-human dialogs of tourist information in Singapore from Skype calls that spanned five domains---accommodation, attraction, food, shopping, and transportation. Each tourist and guide tend to be expressed in a series of multiple turns. The guide is defined as the system in this paper. We transform raw data into examples that fit our experiments. Each example includes an user utterance and its associated slot tags in IOB format~\cite{hakkani2016multi}, user intents, and responsive system actions. Labels of system actions are defined as the concatenation of categories and attributes of speech acts, e.g. $\mathtt{QST\_WHEN}$. $\mathtt{NULL}$ is added as a waiting response from guides when they are expressed in multiple turns. The consecutive guide actions in response to a single tourist utterance is merged as multiple labels. The whole corpus is split into train/dev/test (see Table~\ref{tab:dstc}). Unseen tokens such as words, user intents, slot tags, and system actions in the dev/test set are categorized as $\mathtt{UNK}$.

\subsection{Evaluation Results}
\begin{figure}[htbp]
\includegraphics[width=.9\linewidth]{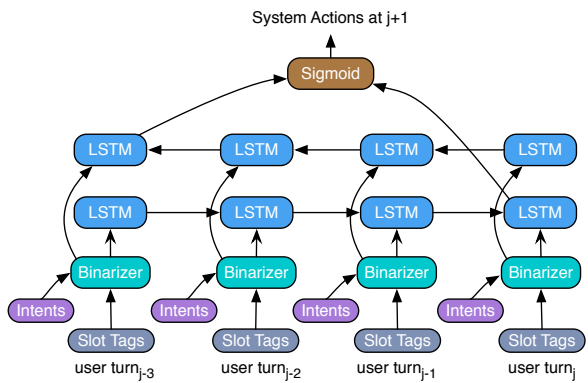}
\caption{biLSTM-based SAP model}
\label{fig:sap_blstm}
\vspace{-2mm}
\end{figure}

We compare our proposed joint model with following models in three tasks: slot filling, intent prediction and SAP\@.
\begin{itemize}
	\item Baseline (CRF+SVMs): NLU and SAP are trained separately, followed by being pipelined for testing. CRF is used to train slot filling model with lexical feature of words; one-vs-all SVMs with linear kernel (LinearSVMs) is used to train intent model with bag-of-words features of user utterances; SAP utilizes LinearSVMs with features of one-hot vectors of aggregated user slot tags and intents. Decision thresholds for intent model and SAP are 0.225 and 0.162.
	\vspace{-1.5mm}
	\item Pipeline (biLSTMs): NLU in Fig.~\ref{fig:nlu_blstm} and SAP in Fig.~\ref{fig:sap_blstm} are separately trained, followed by being pipelined for testing. Best decision thresholds for intent model and SAP model are 0.391 and 0.064.
	\vspace{-1.5mm}
	\item Oracle-SAP (SVMs): The inputs of SAP are clean slot tags and intents annotated by human experts; LinearSVMs is used for training and testing SAP\@. Best decision threshold is 0.162.
	\vspace{-1.5mm}
	\item Oracle-SAP (biLSTM): SAP takes as inputs the same to Oracle-SAP but uses biLSTM for training and testing (see Fig.~\ref{fig:sap_blstm}). Best decision threshold is 0.064.
\end{itemize}

Evaluation results of end-to-end models are illustrated in Table~\ref{tab:joint}. Our proposed joint model outperforms all other end-to-end models in frame-level accuracy by a large margin. The joint model and biLSTMs pipeline achieved absolute increase over baseline with 15.03\% and 4.25\%, respectively. Both models beat the SVMs oracle scores. The biLSTMs pipeline model get worse than biLSTM oracle as expected since it transfer the errors from NLU to the SAP model. Nevertheless, the joint model obtains 10.88\% increase than pipeline model and 3.17\% than biLSTM oracle. These promising improvements indicate that joint training can mitigate the downside of pipeline model in that the hidden outputs from a history of NLU units capture highly more expressive feature representations than the conventional aggregation of user intents and slot tags. In comparison of these two oracle models, the large improvement (12.02\%) for biLSTM model indicates that the contextual user turns make significant contribution to system action prediction. In real human interaction scenarios, frame-level metrics are far more important than token-level ones especially for these multi-label classification tasks since predicting precise number of labels is more challenging.

Evaluation results of NLU models that are frozen as independent models are illustrated in Table~\ref{tab:nlu}. Baseline using CRF and SVMs still maintains a strong frame-level accuracy with 33.13\%, however, biLSTM models taken from pipeline and joint model achieve better increase 3.25\% and 4.25\%, respectively. This observation indicates that joint training with two tasks of slot filling and intent prediction captures implicit knowledge underlying the shared user utterances, while another supervised signal from system actions is capable of refining the biLSTM based model by backpropagating the associated error gradients. Best accuracy at frame-level for slot filling task is obtained by traditional CRF baseline with only lexical features of words, and our biLSTM models fall behind with absolute decrease 0.47\% and 0.82\%. Best frame accuracy for intent prediction task is achieved by our proposed model with 5.21\% improvement.

\section{Conclusion}\label{sec:conclusion}
We proposed an end-to-end deep recurrent neural network with limited contextual dialog memory that can be jointly trained by three supervised signals of user slot filling, intent prediction and system action prediction. Experiments on multi-domain human-human dialogs demonstrated that our proposed model expressed superb advantages in natural language understanding and dialog manager. It achieved better frame-level accuracy significantly than the state of the art that pipelines separate models of NLU and SAP together. The promising performance illustrated that contextual dialog memory made significant contribution to dialog manager, and highly expressive feature representations beyond conventional aggregation of slot tags and intents could be captured in our joint model such that the affects of noisy output from NLU were mitigated. Extra supervised signal from system actions is capable of refining NLU model by backpropagating.

\vfill
\pagebreak

\setlength{\bibsep}{0.75ex}
\small
\bibliographystyle{IEEEbib}
\bibliography{xuesong_refs_slu}

\end{document}